\documentclass[sigconf]{acmart}

\usepackage{stfloats}
\usepackage{graphicx}
\usepackage{subfigure}
\usepackage{amsfonts}
\usepackage{bbm}
\usepackage{amsmath}
\usepackage{cases}
\usepackage{multirow}
\usepackage{multicol}
\usepackage{diagbox}

\usepackage[noend]{algpseudocode}
\usepackage[ruled,vlined, linesnumbered]{algorithm2e}
\usepackage{tikz}
\usepackage{booktabs}
\usetikzlibrary{positioning}
\usepackage{ragged2e}
\usepackage{color,soul}
\usepackage{array}
\usepackage{chngpage}
\usepackage{caption}
\usepackage{enumitem}

\DeclareMathOperator*{\argmax}{arg\,max}

\DeclareMathOperator*{\onehot}{OneHot}
\DeclareMathOperator*{\dense}{Dense}

\usepackage{titlesec}
\titlespacing*{\section}{0pt}{-0.2\baselineskip}{0\baselineskip}
\titlespacing*{\subsection}{0pt}{0.1\baselineskip}{0.1\baselineskip}
\titlespacing*{\subsubsection}{0pt}{0.1\baselineskip}{0.1\baselineskip}

\setcopyright{none}
\renewcommand\footnotetextcopyrightpermission[1]{}
\AtBeginDocument{%
  \providecommand\BibTeX{{%
    \normalfont B\kern-0.5em{\scshape i\kern-0.25em b}\kern-0.8em\TeX}}}

\begin{document}

\title{TG-GAN: Continuous-time Temporal Graph Generation with Deep Generative Models}





\author{Liming Zhang}
\email{lzhang22@gmu.edu}
\orcid{0000-0002-8451-4206}
\affiliation{%
  \institution{George Mason University}
  \streetaddress{4400 University Dr}
  \city{Fairfax}
  \state{Virginia, USA}
  \postcode{22030}
}

\author{Liang Zhao}
\email{lzhao9@gmu.edu}
\affiliation{%
  \institution{George Mason University}
  \streetaddress{4400 University Dr}
  \city{Fairfax}
  \state{Virginia, USA}
  \postcode{22030}
}

\author{Shan Qin}
\email{qinshan2016@bupt.edu.cn}
\affiliation{%
  \institution{Beijing University of Posts and Telecommunications}
  \streetaddress{10 Xitucheng Road}
  \city{Haidian District}
  \state{Beijing, China}
  \postcode{100876}
}

\author{Dieter Pfoser}
\email{dpfoser@gmu.edu}
\affiliation{%
  \institution{George Mason University}
  \streetaddress{4400 University Dr}
  \city{Fairfax}
  \state{Virginia, USA}
  \postcode{22030}
}

\renewcommand{\shortauthors}{Zhang et al.}

\begin{abstract}
  The recent deep generative models for static graphs that are now being actively developed have achieved significant success in areas such as molecule design. However, many real-world problems involve temporal graphs whose topology and attribute values evolve dynamically over time, including important applications such as protein folding, human mobility networks, and social network growth. As yet, deep generative models for temporal graphs are not yet well understood and existing techniques for static graphs are not adequate for temporal graphs since they cannot 1) encode and decode continuously-varying graph topology chronologically, 2) enforce validity via temporal constraints, or 3) ensure efficiency for information-lossless temporal resolution. To address these challenges, we propose a new model, called ``Temporal Graph Generative Adversarial Network'' (TG-GAN) for continuous-time temporal graph generation, by modeling the deep generative process for truncated temporal random walks and their compositions. Specifically, we first propose a novel temporal graph generator that jointly model truncated edge sequences, time budgets, and node attributes, with novel activation functions that enforce temporal validity constraints under recurrent architecture. In addition, a new temporal graph discriminator is proposed, which combines time and node encoding operations over a recurrent architecture to distinguish the generated sequences from the real ones sampled by a newly-developed truncated temporal random walk sampler. Extensive experiments on both synthetic and real-world datasets demonstrate TG-GAN significantly outperforms the comparison methods in efficiency and effectiveness.
\end{abstract}

\maketitle

\section{Introduction}
The domain of generative models for graphs has a long history and over decades many \textit{prescribed} models such as random graphs \cite{erdds1959random} and stochastic blockmodels \cite{goldenberg2010survey} have been proposed as the network generation principles. 
These principles are typically predefined by human heuristics and prior knowledge, which effectively abstract the high-dimension problems down to manageable scale.
Such methods usually fit well towards the properties that have been covered by the predefined principles, but not on those have not been covered.
However, in many domains, the network properties and generation principles are largely unknown yet. 
In recent years, deep generative models for graphs such as GraphRNN \cite{you2018graphrnn} and GraphVAE \cite{simonovsky2018graphvae} have started to achieve significant success over the traditional prescribed ones in more and more applications on static graphs such as molecule design \cite{simonovsky2018graphvae}, thanks to their high expressiveness in learning underlying complex principles directly from the data in end-to-end fashion without handcrafted rules.

\begin{figure}
    \centering
    \includegraphics[width=\columnwidth, trim={0cm 2mm 0cm 0cm},clip]{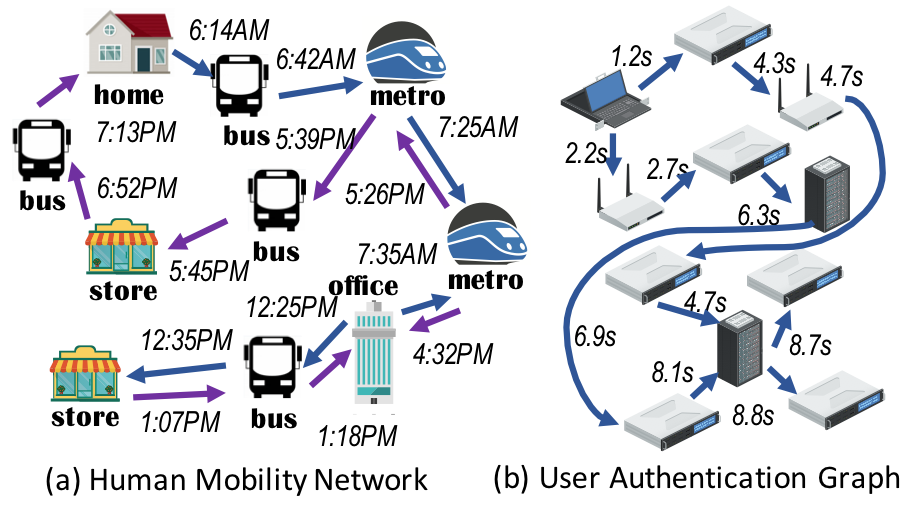}
    \caption{Real-world temporal graphs: a) a transport graph capturing urban movement with destinations of home, office, stores,metro, and bus. Arrows indicate trips between locations at given times. b) a user authentication graph to model cyber behavior. A user uses a laptop to remotely connect different desktops, gate servers, production servers, and data servers chronologically.}
    \label{fig:introduction}
\end{figure}

However, many real-world graphs are actually \emph{temporal graphs} that dynamically evolve over time. Figure \ref{fig:introduction} illustrates a toy example of a mobility network of a person, where each node is a spatial location. A temporal edge between two nodes denote a travel movement at some time. All the temporal edges and nodes in a time period (e.g., a day)  forms a continuous-time temporal graph that reflects whether she is a commuter, traveler, or other behavior type. Similarly there are many other examples such as that in Figure \ref{fig:introduction}(b) which shows an user authentication graph that reflects behavior of an administrator in an enterprise computer network. Here each node is a computer and each temporal edge is the ``login'' behavior from one computer to another. Modeling temporal graphs have significant meaning in network behavior synthesis and intervention, brain network simulation, and protein folding. Given the underlying complex unknown network process, this paper is interested in highly expressive deep models that can distill and model the underlying generative process for generating temporal graphs. This topic goes beyond related domains such as representation learning for dynamic graphs \cite{goyal2018dyngem} and temporal link prediction \cite{yang2019advanced, nguyen2018dynamic}.

However, deep generative models for temporal graphs have not been well explored until now, due to several critical technical challenges that prevent current techniques for static graphs to be easily applied: 
\textit{1) Difficulty in ensuring temporal validity of the generated graphs.} The connectivity patterns of temporal graphs can be highly different from those in static ones, due to the involvement of temporal dimension. For example, in temporal graph, if we want to pass a message from a node $a$ to node $c$ via node $b$, we require not only the existence of edges from $a$ to $b$ and $b$ to $c$, but also that the edge from $b$ to $c$ cannot disappear before the birth of the edge from $a$ to $b$. There are many such types of temporal validity requirements in temporal graphs, which cannot been considered or ensured by current static graph generation techniques.
\textit{2) Inefficiency in data representation in both topological and temporal dimensions.} In temporal graphs, temporal edges usually exists with limited time spans. This results in a highly sparse representation in a joint topological and temporal high dimensional space, especially for graph representation such as graph snapshots of adjacency matrices utilized by most of the techniques for static and dynamic graphs. Hence it is extremely memory- and computational-expensive to directly apply current static graph generation methods to sequentially generate sequence of graph snapshots with highly enough temporal resolution to capture dynamics in continuous time.
\textit{3) Difficulty in encoding and decoding operations jointly for graph topological and temporal information} The node and edge patterns on temporal graphs are simultaneously influenced by graph topology and temporal dependency, which also mutually impact each other. The same path in temporal graph may have different meaning if happening with different time, for example in Figure \ref{fig:introduction}, traveling to downtown area in the morning and midnight corresponds to different underlying behaviors. It is difficult to learn underlying distribution that can concisely represent the interaction between graph topology and temporal dependency.


To address these challenges, we propose a new model, called ``Temporal Graph Generative Adversarial Network'' (TG-GAN) for continuous-time temporal graph generation. It treats temporal graph generation as generation and assemble of truncated temporal random walks and hence  the proposed method is very efficient in handling sparse and non-small graphs with continuous-time evolution. We propose a novel temporal graph generator that jointly model truncated edge sequences via novel recurrent-structured model that enforces temporal validity constraints. We also propose new encoder and decoder for modeling continuous time information and discrete node and edge information. 
The major contributions of this paper can be summarized as follows:
\begin{itemize}[nolistsep,leftmargin=*]
\item \textbf{Propose a novel deep generative framework for temporal graph generation}. The proposed framework can efficiently and effectively learn the underlying distribution of continuous-time temporal graphs. It generates graphs with time-varying node features while ensure temporal validity.

\item \textbf{Develop a new truncated temporal random walk generator}. Encoders and decoders for both continuous-valued time information and discrete-value node information have been proposed. New time budgeting strategies and activation functions have been proposed to ensure temporal validity.

\item \textbf{Propose a new truncated temporal graph discriminator.} A new truncated temporal random walk sampler with temporal jumps is proposed to obtain real samples from observed temporal graphs. A recurrent-architecture based discriminator is developed to jointly examine the sequence patterns and time validity.

\item \textbf{Conduct extensive experiments on both synthetic and real-world data} The results demonstrate that the proposed TG-GAN is capable of generating temporal graphs close to real graphs. It significantly outperforms other models in several metrics by two order of magnitude with outstanding scalability.
\end{itemize}


\section{Related Work}
\label{sec:related_work}
\textbf{Temporal graph generation} 
Graph generative models have important applications in many domains. Conventional methods have been proposed based on a \textit{prescribed} structural assumptions like probabilistic models \cite{barabasi1999mean}, configuration models \cite{bender1978asymptotic}, and stochastic block models \cite{xu2014dynamic}. 
These \textit{prescribed} generative approaches capture some predefined properties of a graph, e.g., degree distribution, community structure, clustering patterns, etc. 
Relevant extensions for temporal graphs are based on these prescribed models and also include models with new designs. 
Stochastic block transition models \cite{yang2011detecting, xu2015stochastic} use a Hidden Markov along with Stochastic-Block-Model (SBM) includin other variant like Temporal Stochastic Block Model (TSBM) \cite{corneli2016exact}. 
A critical limitation of these \textit{prescribed} models is that model assumptions can easily be violated in very big graph datasets, like heavily-tailed degree distributions. 
Some recent works present surprising properties of real-world large graph that deem \textit{prescribed} models to be insufficient. 
We point the reader to comprehensive surveys such as \cite{holme2015modern, rozenshtein2019mining}.

\textbf{Deep generative models for graphs}
Deep graph generation is based on the concept of unsupervised learning. Existing work is based on deep structures like adverversarial network over random walks (NetGAN \cite{bojchevski2018netgan}), variational autoencoders (GraphVAE \cite{simonovsky2018graphvae}), and recurrent networks (GraphRNN \cite{you2018graphrnn}). 
GraphVAE \cite{simonovsky2018graphvae} is a new and first-of-its-kind variational autoencoder for whole graph generation, though it typically only handles very small graphs and cannot scale well to large graphs in both memory and runtime. 
GraphRNN \cite{you2018graphrnn} models a graph as a sequence of node generation and edge generation that can be learned by autoregressive models. It achieves a much better performance and scalability than GraphVAE.
NetGAN \cite{bojchevski2018netgan} follow GAN model \cite{goodfellow2014generative}, and use a generator to generate synthetic random walks while discriminates synthetic walks from real random walks sampled from a real graph.
Typically these approaches can generate a synthetic static graph of good quality. Although a popular generative approach, our literature review has not revealed any deep network models for temporal graph generation.

\textbf{Random walk}
For static graphs, random walk is a powerful and systematically examined representation. The random walk is a diffusion model and its dynamicity provides fundamental hints to the understanding of a whole class of diffusive processes in graphs. 
Temporal graphs are basically time-dependent time-constraint, so temporal random walks capture these dynamic aspects \cite{hoffmann2013random}. For example, continuous-time networks are gradually activated by sequential link connections with start and end times \cite{nguyen2018dynamic}. It will not be validated to activate certain sub-graphs as time evolves. Another reason to use temporal random walk over adjacency matrix representation is its linear running time $O(l)$ with respect to the length of walks. This makes it applicable for very large graphs.
Temporal random walk is used to learn representation in DeepWalk \cite{perozzi2014deepwalk}, dynamic network embeddings \cite{nguyen2018dynamic}, and dynamic representation learning \cite{sajjad2019efficient}. Those contributions cover continuous time representation and discrete time representation with a focus on either node or edge representation.
However, to the best of our knowledge, no existing work try to learn the latent generative distribution for temporal graph generation via temporal random walk.

\section{Problem Definition}
\label{sec:problem}
A temporal graph is a directed graph $G=(V, E, T)$, where $V$ is the set of nodes and $E$ is the set of temporal edges, and $T = [0, t_{end}], \forall \,\, t_{end} \subseteq \mathbb{R}^{+}$ is the time span of the temporal graph. Given two nodes $v_i\in V$ and $v_j\in V$, a temporal edge $e_i$ is denoted as $e_i={u_i,v_i,t_i}\in E$ at time $t_i$, where $u_i$ is the starting node while $v_i$ is the end node. It can also be equivalently denoted as a reversed-timestamp version such that $\bar e_i={u_i,v_i,t_{end}-t_i}\in E$ where $t_{end}-t_i$ is called the \emph{time budget} at the edge $\bar e_i$. Notice that time budget is the residual time left after current temporal edge happens. A temporal random walk is defined as $s=\{e_1,e_2,\cdots, e_{L_s}\}=\{(u_1,v_1,t_1),(u_2,v_2,t_2),\cdots, (u_{L_s},v_{L_s},t_{L_s})\}$, where $L_s$ is the length of this walk and $v_i=u_{i+1}$. An equivalent denotation is $s=\{\bar e_1,\bar e_2,\cdots, \bar e_{L_s}\}$ where each $\bar e_i$ is defined with reversed timestamp such that  $\bar e_i=(u_i,v_i,t_{end}-t_i)$.


The problem of deep generative modeling for temporal graphs is to learn an underlying distribution $p(G)$ of temporal graphs, such that $G\sim p(G)$, namely each temporal graph is a sample from this distribution. Therefore, the training process is to learn $p(G)$ such that the generated graphs minimize its divergence $L()$ (e.g., adversarial loss  and reconstruction loss \cite{goodfellow2014generative}) to the observed samples, under possible temporal and topological constraints. 
Note that different samples of temporal graphs share the same time span while they could have different numbers of nodes and edges.

\begin{figure}[!t]
    \centering
    \includegraphics[width=\linewidth, trim={0.5cm 0cm 1.5cm 0.5cm},clip]{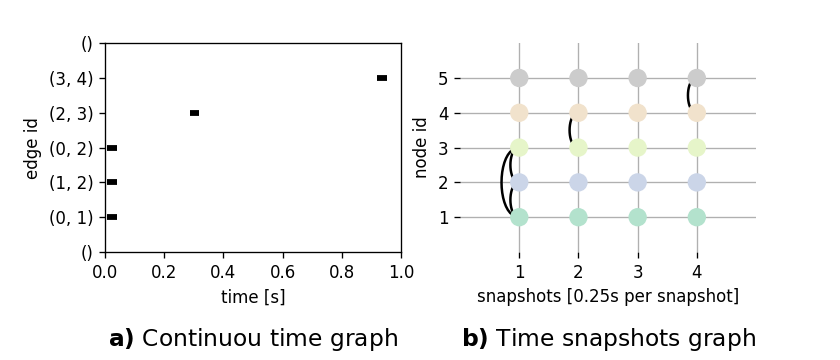}
    \caption{A toy graph in: \textit{a)} continuous time graph with non information loss, length of a line is an edge's contact time; and \textit{b)} discrete-time snapshot graph with information loss}
    \label{fig:example}
\end{figure}
\vspace{6pt}

Since the underlying distribution of an observed temporal graph is unknown, it is prohibitive to directly utilize traditional prescribed models such as temporal random graph and stochastic block models, which require predefined distributions. Therefore, this paper focuses on more powerful models that are sufficiently expressive to learn sophisticated distribution patterns of graph-structured data, by extending the deep generative models for static graphs to temporal graphs. However, though compelling, this research has the following challenges: 1) avoid graphlets (aka. adjacency matrix) to allow for efficient computation for very large graphs; 2) extremely dynamic distribution with temporal hard constraints; and 3) the promise of permutation invariance of nodes like all graph generation models.

\section{Deep Generative Models for Temporal Graph Generation}
\label{sec:method}
To model and generate temporal graphs and address the above challenges, we propose a new model called ``Temporal Graph Generative Adversarial Network'' (TG-GAN). In the following, we present the overall TG-GAN framework and a brief complexity analysis, introduce a novel temporal generator and describe the temporal discriminator and other training strategies.

\subsection{TG-GAN framework}
\label{sec:framework}

\begin{figure*}[!t]
    \centering
    \includegraphics[width=\textwidth, trim={3cm 5.3cm 2.5cm 9.3cm},clip]{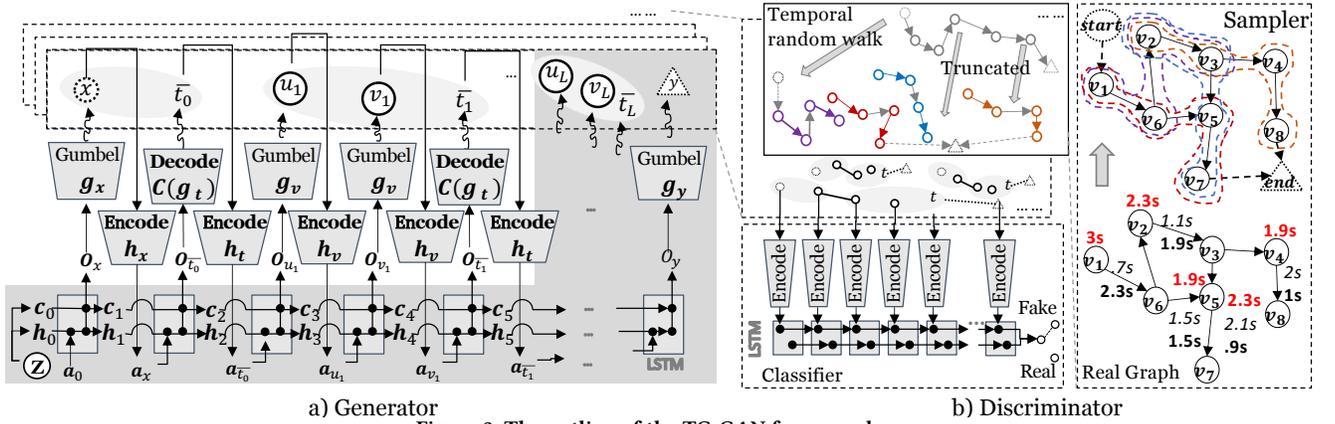}
    \caption{The outline of the TG-GAN framework.}
    \label{fig:gan_model}
\end{figure*}
Overall, the advantages of TG-GAN 
are 1) linear complexity independent of the number of nodes; 2) directly modeling temporal dependency constraints; 3) solving graph generation through sequential models that do not have a permutation requirement.

A temporal graph can be considered as a distribution of a sampled temporal random walks. 
In this paper, we propose TG-GAN, a \textbf{T}emporal \textbf{G}raph generation framework based on \textbf{G}enerative \textbf{A}dversarial \textbf{N}etworks, which captures the topological and temporal patterns of temporal graphs by learning a distribution of a temporal random walk. More concretely, TG-GAN consists of two parts, which are a temporal random walk generator $\mathcal{G}$ and a temporal random walk discriminator $\mathcal{D}$. Generator $\mathcal{G}$ defines an implicit probabilistic model for generating a set of temporal random walks: $\mathcal{G}(\pmb{z})$ that can assemble a temporal graph, where $\pmb{z}\sim p(\pmb{z})$ follows a trivial distribution such as isotropic Gaussian. Note that the generated temporal random walk $\mathcal{G}(\pmb{z})$ must satisfy the temporal constraint. All its valid solution is defined as $\mathcal{C}$. Additionally, the Discriminator $\mathcal{D}$ is developed to discriminate whether any temporal random walk is a ``real'' one from the desired temporal graph or not. Overall, the objective function of TG-GAN with temporal constraints is:
\begin{align}
\nonumber
\label{eq:objfun}
    \min_{\mathcal{G}}\max_{\mathcal{D}} \,\, & \left[ \mathbb{E}_{s \sim p_{data}(s)}[ log(\mathcal{D}(s) ] + \mathbb{E}_{\pmb{z} \sim p(\pmb{z})}[ log(1 - \mathcal{D}(\mathcal{G}(\pmb{z})) ] \right] \\
    \ \ \ \ \ \ \ & \ \ \ \ \ \ \ \ \ s.t. \,\,   \mathcal{G}(\pmb{z}) \in \mathcal{C}, \forall \pmb{z}\sim p(\pmb{z})
\end{align}
where $s$ is a real temporal random walk sampled from observed temporal graphs. To successfully solve the training objective in Equation~\eqref{eq:objfun} for TG-GAN required, we need to address the three key components: 1) Temporal random walks typically are variable-length and are sophisticated and have different characteristics that are more sophisticated than conventional random walk. For example, the edges are distinct and the start and end is dependent on the start time and end times of the temporal graph as well as its topology. Hence, temporal random walk inevitably have variable lengths. So how to effectively generate such types of walks is challenging and will be shown in Section \ref{sec:truncated}. 
2) Simultaneous generation of both continuous and discrete variables. For each temporal edge, we need to jointly generate both discrete nodes and continuous time, and at the same time consider their co-evolving patterns. Section \ref{sec:decode_encode} gives details on how to achieve this. 3) The generated walks must satisfy temporal validity constraints. For each temporal random walk $s=\{e_1,e_2,\cdots, e_{L_s}\}=\{(u_1,v_1,t_1),(u_2,v_2,t_2),\cdots, (u_{L_s},v_{L_s},t_{L_s})\}$, we have a) $t_1\ge 0$, b) $t_{L_s}\le t_{end}$, and c) $t_{i}\le t_{j},\forall i<j$. However, the proposed method needs to satisfy the above constraints while maintaining the efficiency of the generating process. A non-trivial exercise as shown in Section \ref{sec:decode_encode}. 
In this paper, the discriminator and generator directly uses the WGAN framework \cite{arjovsky2017wasserstein} to measure the similarity between the distribution of generated truncated temporal random walks and that of actual ones. \textit{Autodiff} with $Adam$ optimizer is used to optimize our model.

\textbf{Complexity analysis:} TG-GAN is $O(L_s)$ because of non-parallel generation of a whole sequence, where $L_s$ is maximal length of all temporal random walks. Memory complexity is $O(|V| \cdot L)$ for storing logit vectors of sampling nodes, where $L$ is the length of \textit{truncated temporal random walks}. Hence, our proposed model is highly efficient in handling large graphs without any information loss. This elucidates the obvious advantage of our temporal random walk based method over other potential strategies using snapshots and which require at least $O(|V|^2\cdot T)$ but still with information loss, where $T$ is the number of time snapshots.

\subsection{Generator}
\label{sec:gen}
We introduce the general architecture of the generator for the truncated temporal random walk generation, and detail the operations for sampling temporal edges.

\subsubsection{\textbf{Truncated temporal random walks with time budget}}.
\label{sec:truncated}
For continuous-time temporal graphs, the temporal granularity can be infinitely small and the lengths of temporal random walks vary greatly, depending on a temporal process constrained by the start and end time. Such phenomena raise very serious challenges wrt. sequence generation methods such as Recurrent Neural Networks (RNN) \cite{lecun2015deep}. First, although modern operations (e.g., LSTM \cite{lecun2015deep} and Transformer \cite{vaswani2017attention}) and activation functions (e.g., ReLU \cite{lecun2015deep}) have been proposed to remedy the gradient vanishing issues in sequence learning, it is still highly challenging for backpropagation-based methods to learn very long sequences \cite{lecun2015deep}. Something that applies to our problem. Second, directly learning variable-length temporal edge sequences requires sequence learning methods to jointly learn not only the distribution of nodes and timestamps, but also the distribution of sequence lengths using the recurrent module. Such multi-modal sophisticated tasks easily overwhelm the model capability and lead to a failed learning task.

To address the challenge, we propose to learn whole (variable-length) temporal random walks as concatenations of multiple truncated temporal random walks under a time budget and of shorter length. This is illustrated in upper right corner of Figure \ref{fig:gan_model}(b). When learning the whole sequence (from \textit{start} circle node to \textit{end} triangle node), the proposed sequential learning method only learns the subsequence (colored ones) with a length equal to or smaller than $L$ with a little additional profile information for later-on concatenation operation. More formally, a truncated temporal random walk is defined as the follows.

\begin{definition}[Truncated Temporal Random Walks]
A \textbf{truncated temporal random walk}  is defined as a sequence \\ $\bar s=\{c,\bar e_1,\bar e_2, \cdots, \bar e_{L}\}$ happens within the temporal range of $t_0\ge 0$ and $t_{end}$. $\bar s$ consists of its profile $c$ and the temporal edges $\bar e_i, i=1,\cdots,L$, where $L$ is equal to or less than a threshold defined as the maximal length of a truncated temporal random walk. Here the profile $c=(x,y,\bar t_0)$ includes $x\in\{0,1\}$ denoting whether $\bar s$ is the initial ($=1$) or not ($=0$) in a temporal random walk, while $y\in\{0,1\}$ denoting whether it is the end ($=1$) or not ($=0$). $\bar t_0=t_{end}-t_0$ is defined as the \textbf{time budget} of $\bar s$, where $t_0 = t_{end}, \forall x = 1$ for initial truncated walks, or $\bar t_0 = \bar t_{L_{s-1}}, \forall x = 0$, and $\bar t_{L_{s-1}}$ is time budget of previous temporal edge in a whole walks.
\end{definition}

\begin{figure}[!]
    \centering
    \includegraphics[width=\linewidth, trim={2cm 4.5cm 8cm 4cm},clip]{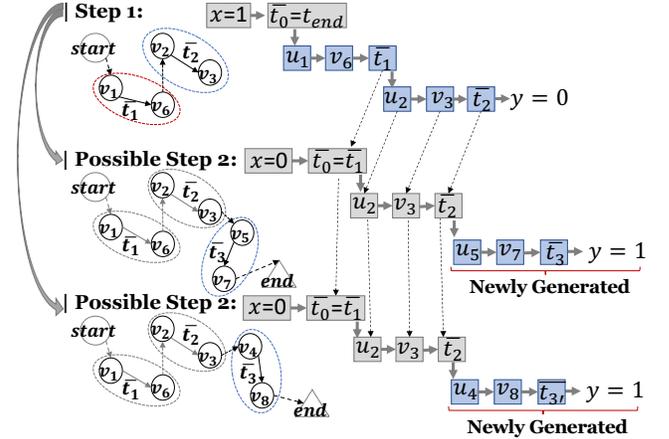}
    \caption{A toy example for non-parallel generation of temporal random walks through truncated temporal random walks. \textit{Step 1)}: generate two temporal edges (blue squares) with $x=1, \bar t_0=t_{end}$ as inputs; \textit{Step 2)}: using Step 1 generate only one additional edge and $y$ profile. Two possible outputs can be achieved with different probabilities so as to capture the randomness in temporal graphs.}
    \label{fig:generate_truncated}
\end{figure}

For example, in Figure \ref{fig:gan_model} (b), a temporal random walk with a length of 3 has been concatenated as two truncated temporal random walks, where the first one starting at $u_1$ is shown as a purple dashed line, hence, its profile time budget is $\bar t_0=t_{end}=3s$ and its starting flag is$x=1$. Raw time stamps are shown in italic. The random walk has two edges, which are ($u_1$, $v_6$, 2.3s) and ($u_2$, $v_3$, 1.9s). since it is not the end of the temporal random walk, its end flag is $y=0$. 2.3s and 1.9s are the remaining time budget after these edges are created at time 0.7s and 1.1s. The second truncated temporal random walk (blue dashed line) starts at ($u_2$, $v_3$, 1.9s) with a profile of $x=0$, and $\bar t_0 = 2.3s$, here 2.3s comes from previous edge ($u_1$, $v_6$, 2.3s). The second edge is ($u_5$, $v_7$, 1.5s), and finally, $y=1$ for this walk. Notice that the second edge could also be ($u_4$, $v_8$, 1s) with $y=1$ (orange line). However, if the first truncated sequence is ($u_1$, $v_6$, 2.3s), ($v_5$, $v_7$, 1.5s), the other edges would not be reachable since the whole walk ends with $y=1$. This is an example of variable-length temporal random walks aka. a time-dependent graph topology. The corresponding detailed generation of these walks is also found in Figure \ref{fig:generate_truncated}.
\begin{equation}
\label{equ:lstm}
\begin{split}
    \pmb{a}_0 = \pmb{0},\,\,\,\, \pmb{m}_0 = h_0(\pmb{z}),\,\,\,\, \pmb{z} \sim U(\pmb{0}, \pmb{1}) \\
    \pmb{a}_{x} = h_x(x),\,\,\,\, x = g_x(\pmb{o}_{1}),\,\,\,\, (\pmb{m}_1, \pmb{o}_{1}) = f_{\theta}(\pmb{m}_0, \pmb{a}_0) \\
    \pmb{a}_{\bar t_0} = h_t({\bar t_0}),\,\,\,\, {\bar t_0} = \mathcal{C}(g_t(\pmb{o}_{2})),\,\,\,\, (\pmb{m}_2, \pmb{o}_{2}) = f_{\theta}(\pmb{m}_1, \pmb{a}_{x}) \\
    \pmb{a}_{u_1} = h_v(u_1),\,\,\,\, u_1 = g_v(\pmb{o}_{3}),\,\,\,\, (\pmb{m}_3, \pmb{o}_{3}) = f_{\theta}(\pmb{m}_2, \pmb{a}_{\bar t_0}) \\
    \pmb{a}_{v_1} = h_v(v_1),\,\,\,\, v_1 = g_v(\pmb{o}_{4}),\,\,\,\, (\pmb{m}_4, \pmb{o}_{4}) = f_{\theta}(\pmb{m}_3, \pmb{a}_{u_1}) \\
    \pmb{a}_{\bar t_1} = h_t({\bar t_1}),\,\,\,\, {\bar t_1} = \mathcal{C}(g_t(\pmb{o}_{5})),\,\,\,\, (\pmb{m}_5, \pmb{o}_{5}) = f_{\theta}(\pmb{m}_4, \pmb{a}_{v_1}) \\
    \dots \\
    y = g_y(\pmb{o}_{3l+3}),\,\,\,\, (\pmb{m}_{3l+3}, \pmb{o}_{3l+3}) = f_{\theta}(\pmb{m}_{3l+2}, \pmb{a}_{3l+2}) \\
\end{split}
\end{equation}

To learn the generative process of temporal random walks, a novel recurrent-architecture-based sequential model (abstracted operations in Equation \ref{equ:lstm}) is proposed here to characterize and concatenate the truncated temporal random walks that compose them. 
This sequential model learns $x_0$, $\bar t_0$, $\bar e_1$, $\bar e_2$, $\cdots$, $\bar e_{L}$, and $y_0$ sequentially, as shown in Figure \ref{fig:gan_model}. 
The length $L$ is fixed to a small value (e.g., $1 \le l \le 20$) depending on the maximum length of the whole temporal walks. 
With this efficient architecture for smaller sequences, we can still preserve the temporal dependency across different adjacent pieces, since such a dependency has been effectively encoded in each piece's profile of  $x_0, \bar t_0, y_0$. The encoded profile information is absorbed into the first two and the last recurrent units, which can have substantial instructions to the next ones. This addresses the first two challenges mentioned above re. variable-length temporal random walks.
The basic recurrent units shown at the bottom of the generator in Figure \ref{fig:gan_model} utilize an LSTM model. We input a latent code vector $\pmb{z} \in \mathbb{R}^h$ into the first recurrent unit, which can be sampled from some trivial distributions such as a multivariate uniform distribution or Multivariate Gaussian. In Equations \ref{equ:lstm}, $g_y, g_x, g_t, g_v$ being different decoding operations, $h_x, h_t, h_v$ are different encoding operations as detailed in \ref{sec:decode_encode}. Because we focus on LSTM, a memory state $\pmb{m}^{\kappa}$ is composed of a cell state $\pmb{c}^{\kappa}$ and $\pmb{h}^{\kappa}$.

\textbf{Temporal random walk generation using the trained generator:} Once the above generator has been trained, truncated temporal random walks will be generated to compose the whole temporal random walk. However, we found that simply generating truncated temporal random walks and then arbitrarily matching and chronologically concatenating them is not effective, since it does not preserve the temporal dependency of two consecutive truncated walks. To address this, when generating each temporal edge $\bar e_i$, we need to maintain its conditional dependency $p(\bar e_i, y_0|\bar e_{i-1},\bar e_{i-2},...\bar e_{1},x_0,t_0)$ over all the historical edges in this temporal random walk. Therefore, as shown in Figure \ref{fig:generate_truncated}, we propose to first generate an initial truncated temporal random walk and then incrementally append it with one additional temporal edge $\bar e_i$ as well as an end status flag $y_0$. This process re-uses the same generator and ends when $y = 1$ is found.

\subsubsection{\textbf{Decoding and Encoding Operations for Time information under Temporal Constraints}}.
\label{sec:decode_encode}

To generate valid time information under temporal constraints, the output $\pmb{o}^{\kappa}$ from the recurrent unit of each time step needs to be decoded into a real value under validity constraints. The critical challenges include learning the arbitrary underlying distribution of time while ensuring the temporal validity of temporal graphs. For, we propose the following.

\textit{\textbf{Non-parametric time distribution inference}}
The time of each temporal edge is assumed to be sampled from underlying distributions, which could be simply assumed as a Gaussian, Gamma, or Beta distribution. Then a dense layer is established to map $\pmb{o}^{\kappa}$ to a sufficient statistic. For example, assuming a Gaussian distribution, then two neural networks can be built to map $\pmb{o}^{\kappa}$ to the mean $\mu$ and $\sigma$ of the distribution, respectively. 
However, when training the model, since backpropagation cannot handle uncertain parameters, the re-parameterization trick will be adopted to move the non-differentiable sampling operations from the sufficient statistics parameters (e.g., mean and variance of Gaussian) to a unit Gaussian $\mathcal{N}(0, 1)$ (cf. \cite{kingma2013auto}). 
Such distributions are called \textit{parametric} distributions \cite{wasserman2013all} (Equation \ref{eq:repara} shows an example for a Gaussian distribution).
\begin{equation}
\label{eq:repara}
\begin{split}
    \mu = \dense(\pmb{o}^{\kappa}), \,\,\,\, \sigma = \dense(\pmb{o}^{\kappa}) \\
    t_i' \sim \mathcal{N}(0, 1), \,\,\,\, t_i = \mu + \sigma t_i'
\end{split}
\end{equation}

\begin{figure}[!t]
    \centering
    \includegraphics[width=\linewidth, trim={7.5cm 8.2cm 6.5cm 9cm},clip]{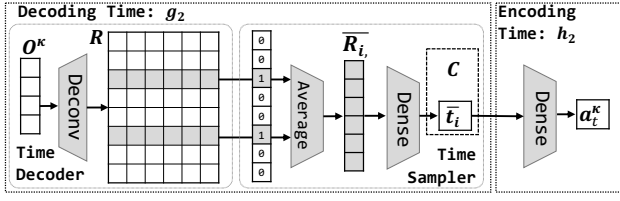}
    \caption{Deep Temporal Random Sampler for non-parametric distributions of continuous time}
    \label{fig:shoot}
\end{figure}

In many situations, the real time distribution cannot be simply fit to an existing simple parametric distribution since they are \textit{non-parametric}. 
For example, the time of a trip from one location to another might be specific to each commuter when consider traffic and job requirements and we cannot simply assume a Gaussian or Gamma distribution since the underlying distribution is unknown. Therefore, it is highly desired for the model to identify and fit such unknown distributions. Here we propose a novel non-parametric time distribution inference method named \textbf{Deep Temporal Random Sampler} as illustrated in Figure \ref{fig:shoot}. 
It consists of two modules: 
1) a time decoder, which is a series of deconvolutional layers $Deconv$ that can project an initial variable $\pmb{o}^{\kappa} \in \mathbb{R}^H_o$ to a matrix $\pmb{R} \in \mathbb{R}^{D_1 \times D_2}$ of larger dimensions. 
2) a time sampler, which uniformly selects one or multiple rows $\pmb{R}_{i,}$ from $\pmb{R}$ and then averages the selected ones. This averaged vector $\Bar{\pmb{R}}_{i,}$ is mapped to $\bar t_i$ through another Dense layer, i.e., $\bar t_i = \onehot(\Bar{\pmb{R}}_{i,})$. Finally, there is an encoding operation, another Dense layer that maps the generated time back to hidden vectors by $\pmb{a}^{\kappa}_t = \dense(\bar t_i)$, which is then input into the recurrent unit as shown in Figure \ref{fig:shoot} and Equation \ref{eq:deep-shoot}. 

\textit{\textbf{Temporally-valid activations for ensuring time constraints}}
Different from conventional random walks, temporal random walks require that the temporal edges also satisfy various time constraints depending on specific applications \cite{hoffmann2013random}. 
With the nature of time, the time budget $\bar t_i,\ (i=0,1,\cdots)$ will always be at least non-negative. Moreover, for each truncated temporal random walk, the time budgets of all temporal edges should be no less than 0 and monotonically non-increasing $\bar t_0\ge\bar t_1\ge\bar t_2\ge\cdots\ge 0$. Sometimes the time budget in contiguous edges cannot be equal, such as in human mobility networks, where a user cannot appear in two locations at the same time, $t_i \ne t_{i-1},\ \forall t_i\le t_{end}$. The time values can be normalized to a range between 0 and 1 by scaling $t_{end}$ down to $1$, which can be easily recovered back to the original scale when needed. For ease of analysis, in the following all the time values are normalized ones.

It is usually highly difficult for backpropagation methods to ensure all the above constraints, which are typically non-differentiable and mutual dependent. To address various temporal constraints, we essentially need to provide both lower- and upper-bounds for all the time values that have been sampled. To achieve this, we propose the following time value bounding methods: 1) \textit{Clipping}, which clip the generated time to boundary values if it is outside the range, similar to image generation \cite{gatys2015neural}; 2) \textit{Nested Relu Bounding} to ensure $0\le\bar t_i\le \bar t_{i-1}$, which uses two nested\textit{Relu} functions $\bar t_i =  Relu(\bar t_{i}) - Relu(\bar t_{i} - \bar t_{i-1})$; 3) \textit{Minimax Bounding}, which uses a scaling based on ``min'' and ``max'' values within the sampled set of truncated sequences $\{\bar s\}$ for each training epoch (called a training mini-batch in other literatures). 
It obtains a minimum value from this set first, $min(\{\bar t_i\})$. Then, if $min(\{\bar t_i\}) \leq \epsilon$, then $\bar t_i = \bar t_i - min(\{\bar t_i\})$, otherwise, go to next step. Here, $\epsilon$ is a hyper-parameter with a small value (e.g. $\epsilon < 1e-3$) to prevent a zero value for $\bar t_i$. Next, maximum value of a mini-batch $max(\{\bar t_i\})$ is got. Then, if $max(\{\bar t_i\}) > 1$, then, $t_i = \bar t_i / max(\{\bar t_i\})$, otherwise, nothing is done. This is actually a similar operation like Batch Normalization \cite{ioffe2015batch}.
In Equation \ref{eq:objfun}, different types of constraining operations are represented as the function $\mathcal{C}$.
\begin{equation}
\label{eq:deep-shoot}
\begin{split}
    \pmb{R} = Deconv(\pmb{o}^{\kappa}), \,\,\,\,
    \Bar{\pmb{R}}_{i,} = \frac{1}{n} \sum^n_{i=1} \pmb{R}_{\substack{i \sim Cat(\frac{1}{D_1}), \forall 1 \le i \le D_1}} \\
    \bar t_i =\mathcal{C}\big(\dense(\Bar{\pmb{R}}_{i,})\big), \,\,\,\,
    \pmb{a}^{\kappa}_t = \dense(\bar t_i)
\end{split}
\end{equation}

\noindent\textbf{Decoding and Encoding Operations for Categorical Data}. As shown in Figure \ref{fig:gan_model}(a), categorical data includes nodes, starting flags, and ending flags. Each of them share the same encoding and decoding procedure for each step in the sequence. Take the operations for a node as an example, specifically, first $f_{\theta}$ produces one output vector $\pmb{o}^{\kappa} \in \mathbb{R}^H$ from the recurrent hidden unit (e.g., each LSTM unit). Here $H \ll |V|$ is the dimension of the embedding space of a node from $V$. So $\pmb{o}^{\kappa}$ is a highly concise representation that largely reduces the computing overhead especially for large graphs. Then $\pmb{o}^{\kappa}$ is projected by the function $g_{v, up}(\pmb{o}^{\kappa})$ upscale to another vector $\pmb{q}_v^{\kappa} \in \mathbb{R}^{|V|}$, which is a logit parameter of a categorical distribution for sampling a node $v_{i} \sim Cat(\pmb{q}_v^{\kappa})$. The projection function $g_{v, up}(\pmb{o}^{\kappa})$ is defined as an affine transformation $g_{v, up}(\pmb{o}^{\kappa}) = W_{up} \pmb{o}^{\kappa} + b_{up}$. 
The procedures for starting and ending flags are the same as the above except that the dimension of the decoded value is two, namely $\pmb{q}_x^{\kappa}, \pmb{q}_y^{\kappa} \in \mathbb{R}^2$, and hence the sampling of the starting and ending flags can be denoted as $x \sim Bern(\pmb{q}_x^{\kappa})$ or $y \sim Bern(\pmb{q}_y^{\kappa})$, where $Bern$ denotes the Bernoulli distribution \cite{wasserman2013all}. $x, y, v_i$ are one-hot vectors in this context.

Figure \ref{fig:gan_model}(a) shows the encoding operations for categorical data. Beside memory states, the decoded categorical data (i.e., either node $v_i$ or starting flag $x$) in last time step is also used as input to the LSTM unit. Specifically, to convert categorical data including $x, v_i$, embedding layers are used. The two embedding matrices  $W_{x, down} \in \mathcal{R}^{2 \times H_x}, W_{v, down} \in \mathcal{R}^{|V| \times H_v}, H_v \ll |V|$ are used. $H_v, H_x$ are the dimensions of the embedded vectors. All nodes share the same embedding layer. Embedded vectors are passed to two different dense layers $g_{x, down}, g_{v, down}$ and generate input vectors $\pmb{a}^{\kappa}$ for the next LSTM unit as follows:
\begin{equation}
\begin{split}
\pmb{a}^{\kappa}_v = \dense(W_{v, down}v), \,\,\,\, \pmb{a}^{\kappa}_x = \dense(W_{x, down} x)
\end{split}
\end{equation}

\noindent{\textbf{Gumbel-Max re-parameterization to generate categorical values}}. Sampling  values from categorical (or Bernolli) distributions pose significant challenges for backpropagation training, which inherently requires differentiable objective functions to work. To address this issue, we leverage a reparametrization trick based on Gumbel-Max  \cite{jang2016categorical}
Specifically, we create $v'^{\kappa} = \tanh((\pmb{q}^{\kappa} + \pmb{g}) / \tau)$, 
where $\tau$ is so-called ``temperature'' hyper-parameter. Each value $g_i$ in $\pmb{g}$ is an independent and identically distributed (i.i.d.) sample from standard Gumbel distribution \cite{jang2016categorical}. 
We generate a one-hot representation $v^{\kappa}$, whose $i'$th element is one and all the others are zeros, where $i'=\arg\max_i v'^{\kappa}$. In this way, gradients can be backpropagated through $v'^{\kappa}$. The same approach is also used for $x, y$ sampling. Notice that the larger $\tau$, the more uniformly regulated with more stable gradient flow are the sampled values. The typical approach is to decrease $\tau$ as training continues and we can adopt a decrease strategy similar to \cite{jang2016categorical} (Equations \ref{eq:decode_cat}).
\begin{equation}
\begin{split}
\label{eq:decode_cat}
    \text{for nodes $v_i$, \hspace{3cm}}
    \pmb{q}_v^{\kappa} = W_{v, up} \pmb{o}^{\kappa} + b_{v, up} \\
    v'^{\kappa} = \tanh \big((\pmb{q}^{\kappa}_v + \pmb{g}_v) / \tau \big), 
    v^{\kappa} = \onehot(\argmax v'^{\kappa}) \\
    \text{for start indicator $x$, \hspace{2cm}}
    \pmb{q}_x^{\kappa} = W_{x, up} \pmb{o}^{\kappa} + b_{x, up} \\
    x'^{\kappa} = \tanh \big((\pmb{q}^{\kappa}_x + \pmb{g}_x) / \tau \big),
    x^{\kappa} = \onehot(\argmax x'^{\kappa}) \\
    \text{for end indicator  $y$, \hspace{2cm}}
    \pmb{q}_y^{\kappa} = W_{y, up} \pmb{o}^{\kappa} + b_{y, up} \\
    y'^{\kappa} = \tanh \big((\pmb{q}^{\kappa}_y + \pmb{g}_y) / \tau \big),
    y^{\kappa} = \onehot(\argmax y'^{\kappa}) 
\end{split}
\end{equation}

\subsection{Discriminator via temporal random walks sampler}
\label{sec:disc}

Since TG-GAN iteratively generates and discriminates truncated temporal random walks, the discriminator needs new techniques to sample truncated temporal random walks from real temporal graphs and also sequence classifiers that can classify such truncated temporal edge sequences.

\subsubsection{\textbf{Truncated temporal random walks sampler}}.
\label{sec:sampler}
In the case of a temporal graph, a time-variant graph topology changes dynamically and a variable-length walk sequence can be sampled. If we truncate this variable-length sequences to small fixed-length sequences, there are two challenges:
1) Different sub-sequences of a temporal sequence contain more information, but are more challenging to finding patterns, for most cases, earlier parts are more difficult.
2) Sparse connections of different sub-graphs in a whole graph prevent the learning of global information. In the extreme case, walkers could be stuck in a sub-graph with no out-links to the whole graph. This is referred to as the ``SpaderTrap'' problem in graph data mining \cite{bidoni2014generalization}.

As such, to address the above challenges, we propose a novel sampling method for truncated temporal random walks, which first determines the starting edge of the walk and then samples the next edges sequentially following a temporal process. The proposed truncated temporal random walk sampler is shown in Algorithm \ref{algo:sampler}. Line 1 is to normalize time so that $t_{end} = 1$. Line 2 states the selection of a graph sample, Lines 3-10 are used to sample the profile and walk sequences.
The details of the starting edge sampler $\mathcal{K}$ and next edge sampler $\mathcal{H}$ are elaborated in the following.
\begin{algorithm}[!t]
\small
\caption{Truncated temporal random walks sampler}
\label{algo:sampler}
\DontPrintSemicolon
\SetAlgoLined
\KwData{$E = \{E_d\}, \forall E_d = \{\bar e_i (u_i, v_i, \bar t_i)\}, t_{end}, L$}
\KwResult{a set of truncated sequences: $\{(x, t_0, \bar e_1, \dots, \bar e_L, y)\}$ }
 initialize $\bar t_i \gets \bar t_i / t_{end}$ \;
 sample $d \sim Uniform(1 / |E|)$, and get $E_d$\;
 $\bar e_{i_0}(v_{i_0}, u_{i_0}, \bar t_{i_0}) \sim \mathcal{K}(E_d)$ \;
 \lIf{$i==1$}{
 $x \gets 1$, $\bar t_0 \gets 1$ \textbf{else} $x \gets 0$, $\bar t_0 \gets \bar t_{i-1}$
 }
 $i \gets i_0$ \;
 \While{$i \leq i_0 + L$}{
  $\bar e_{i+1}(u_{i+1}, v_{i+1}, \bar t_{i+1}) \sim \mathcal{H}(\bar e_i)$\;
  $i \gets i+1$
 }
\lIf{$i == |E_d|$}{$y \gets 1$ \textbf{else} $y \gets 0$}
\end{algorithm}

%
%


\textbf{Starting edge sampler $\mathcal{K}$:} Several alternatives could be used, such as a uniform distribution: $p_{\mathcal{K}}(e) =1/|E|$. However, more reasonable ways could be a distribution biased towards the start time, i.e., starting edges that happen earlier have a higher probability. To achieve this, we could leverage a linearly-biased distribution: $p_{\mathcal{K}}(e) =t_i/\sum_{e_i\in E} t_i$ or exponential distribution: $p_{\mathcal{K}}(e) =exp(t_i)/\sum_{e_i\in E}exp(t_i)$.

%
%
%
%
%
\textbf{Descendant edge sampler $\mathcal{H}$ with temporal jumps:} A descendant edge following the current edge could be selected among all its adjacent edges either uniformly or considering time decay. By extending the notion of jumps, we propose temporal jumps, which should help achieve ``smoothness'' in temporal random walks. For example, in human mobility in metro networks can be modeled as a temporal graph in which a traveler could do temporal walks across different metro stations along the transport network (i.e., temporal edges). Temporal jumps occur when a traveler uses intermittently other modes of transport, e.g., walking, taxis, etc. To provide for more robust and flexible modeling, we propose the use of ``teleport temporal edges'' with monotonically-increasing time stamps, by adopting Bayesian prior into temporal graphs.
This teleport temporal edges is a Bayesian prior-enhanced categorical distribution $Cat(p_{\mathcal{H}})$ with a probability of selecting the next edge according to time, is implemented with an exponential time decaying function and a uniform distribution over all the other nodes except the current node.
\begin{align}\nonumber
e_i\sim \mathcal{H}(e_i|\pmb{m}),\ \ m_i= \alpha \left(exp(t_i)/\sum\nolimits_{j = i}^{|E|}exp(t_j) +2 \epsilon /(|V| -1)\right)
\label{eq:jump}
\end{align}
%
where $\alpha$ is a normalization term to make sure all probabilities sum up to $1$, $\epsilon$ is a very small teleport probability over all the other nodes except the current node.

\subsubsection{\textbf{Classifier designs}}.
\label{sec:classifier}
The discriminator $\mathcal D$ is based on the recurrent architecture where the recurrent units could adopt units such as LSTMs. Each input is namely a truncated temporal random walk $\bar s$, which consists of $x$, $\bar t_0$, $\bar e_1$, $\bar e_2$, $\cdots$, $y$ sequentially.  We can directly leverage their encoders introduced in Section~\ref{sec:gen} to encode them and input into each LSTM unit.
After processing the entire temporal edge sequence, the discriminator outputs a single score from the last LSTM unit as the probability of a truncated temporal random walks reflecting an actual walk. \textbf{Training stopping criteria:} TG-GAN uses an early-stopping mechanism that relies on a specific MMD distance (e.g., MMD in Average Degree) to save time for the case of large graphs.
The competitor methods use their default training mechanism.

\section{Experiments}
\label{sec:experiment}
In this section, performances of TG-GAN framework are evaluated using three synthetic datasets and two real-world datasets. Section \ref{sec:exp_setup} introduces the experimental setup. The performance of TG-GAN in terms of Maximum Mean Discrepancy (MMD) in different graph measures is then evaluated against existing deep generative models and a prescribed model in Section \ref{sec:exp_performace}. Finally, the visual qualitative analysis on generated graphs are examined. All the experiments were conducted on a 64-bit machine with a 10-core processor (i9@3.3GHz), 64GB memory, and Nvidia 1080ti GPUs.

\subsection{Experimental Settings}
\label{sec:exp_setup}

\subsubsection{Synthetic datasets}
3 synthetic datasets with increasing complexity are from scale-free random graphs \cite{barabasi1999mean}. As a widely adopted random graph model, a scale-free graph is a static graph whose degree distribution follows a power law and it is constructed by progressively adding nodes to an existing network and introducing links to existing nodes with preferential attachment such that the probability of linking to a given node $i$ is proportional to the number of existing links $k_i$ that node has. We modify this step by adding a time-dependent step (details in supplement \ref{supplement:simulation}. This method was used to generate 3 synthetic datasets with different numbers of nodes, $\{100, 500, 2500\}$. For each dataset, different iterations of simulation (a graph sample), i.e., $\{200, 100, 100\}$, are used.

\subsubsection{Real-world datasets}.
{\it a) User authentication graph} includes the authentication activities of 97 users on their accessible $27$ computers or servers (nodes in graph) in an enterprise computer network during a $485h$ period \cite{akent-2015-enterprise-data}. 
For this work, we choose a single user profile. Each hour is treated as a temporal graph sample. All times are normalized a range of $[0, 1]$.
{\it b) Public transport graph} data capture by farecard records from the Washington D.C. metro system ($91$ stations as graph nodes) includes million of users` trips records. The trip record is in the form <user id, entry station, timestamp, exit station, timestamp>. The dataset captures all trips during a period of three months ($123$ days) from May 2016 to July 2016. Each day is treated as a temporal graph sample. Since metro operations stops at 1am, we shift all the timestamps by one hour to adhere to a $24h$ interval. All timestamp are converted to $[0, 1]$ interval.
All the datasets used are split into a $80\%$ training and $20\%$ test dataset. For the traditional model, which uses considerable main memory, sparser graph samples are used to adhere to memory limits.

\subsubsection{Metrics}
Maximum Mean Discrepancy (MMD) \cite{gretton2007kernel} is chosen to evaluate the distances of graph samples generated from different generative models to real graph samples. MMD is more chosen today (cf. \cite{bounliphone2015test, you2018graphrnn}) instead of Kullback–Leibler divergence and others \cite{lecun2015deep} for similarity of high-dimensional distributions. To ensure a fair comparison, both continuous-time measures and snapshot measures are used. The continuous-time measures include Average Degree, Mean of Average Degree, Group Size, Average Group Size, Mean Coordination Number, Mean Group Number, and Mean Group Duration. The snapshot measures includes Betweenness Centrality, Broadcast Centrality, Bursiness Centrality, Closeness Centrality, Node Temporal Correlation, Receive Centrality, and Temporal Correlation. The lower the MMD value, the better a generative model performs.

\subsubsection{Competing methods}
This experiment utilizes 4 comparison methods: GraphRNN \cite{you2018graphrnn}, NetGAN \cite{bojchevski2018netgan}, GraphVAE \cite{simonovsky2018graphvae}, and a \textit{prescribed} method, Dynamic-Stochastic-Blocks-Model (DSBM) \cite{xu2014dynamic}. We first create snapshots for these methods, train the models, and recover the continuous time from the generated snapshots (Supplement \ref{supplement:revocer}). Necessary parameter-tuning is done first to ensure that TG-GAN performs as expected (Supplement  \ref{supplement:tuning}). 

\subsection{Quantitative performance}
\label{sec:exp_performace}
\begin{figure}[!t]
    \centering
    \includegraphics[width=\linewidth, trim={6cm 4.5cm 3.5cm 4cm},clip]{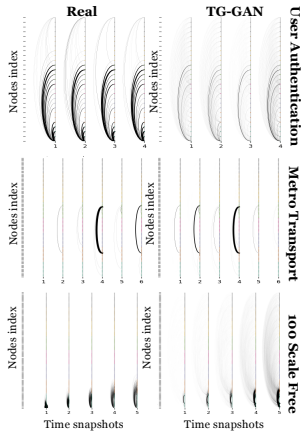}
    \caption{Comparisons of real graphs (left most column) and TG-GAN generated graphs (right column) for different datasets}
    \label{fig:snapshot_viz}
\end{figure}
What follows is a discussion of the performance of TG-GAN in relation to its comparison methods. Given its scalability (or lack thereof) of different methods, GraphVAE is omitted for the 2500-node dataset, and DSBM is omitted for the 500-node and 2500-node datasets. $-$ indicates that programs could not finish running within a reasonable amount of time during our experiments.

\subsubsection{Performance on synthetic datasets}.
Tables \ref{tab:mmd_continuous} and \ref{tab:mmd_discrete} contain MMD distances for different graph measures and the 3 synthetic datasets (indicated as 100, 500, 2500 in the first column). The lower the values are, the better the respective performance. The proposed TG-GAN consistently outperform the other methods. Several metrics, like Average Degree, Average Group Size, Mean Coordination Number achieve a two orders of magnitude improved performance. This shows how much information loss a discrete-time snapshot could have no matter what basic model it uses. Another observation is that TG-GAN, GraphRNN and GraphVAE, generally achieve a lower value than DSBM. It indicates that deep generative methods have a better performance than \textit{prescribed} model. 

For discrete-time snapshot graph measures, the competitive advantage of TG-GAN is not that considerable. In some cases (Betweenness Centrality, Receive Centrality for 100-node graphs), DSBM shows better performance that deep generative methods. TG-GAN outperforms the other two deep generative methods for 100-node graphs. However, the advantage of TG-GAN is not that significant for the 500-node graph. We can see that, TG-GAN still shows the best performance for several measures including Betweenness Centrality, Broadcast Centrality, Receive Centrality, and Temporal Correlation. The lower advantage for larger synthetic graphs could be result of our early-stopping mechanism, which has the potential of better training.

\subsubsection{Performance on real-world datasets}
\begin{figure*}
\scriptsize
\begin{adjustwidth}{0in}{-.5in}
  \begin{minipage}{0.3\linewidth}
      \includegraphics[width=\linewidth, trim={0cm 4cm 15cm 4cm},clip]{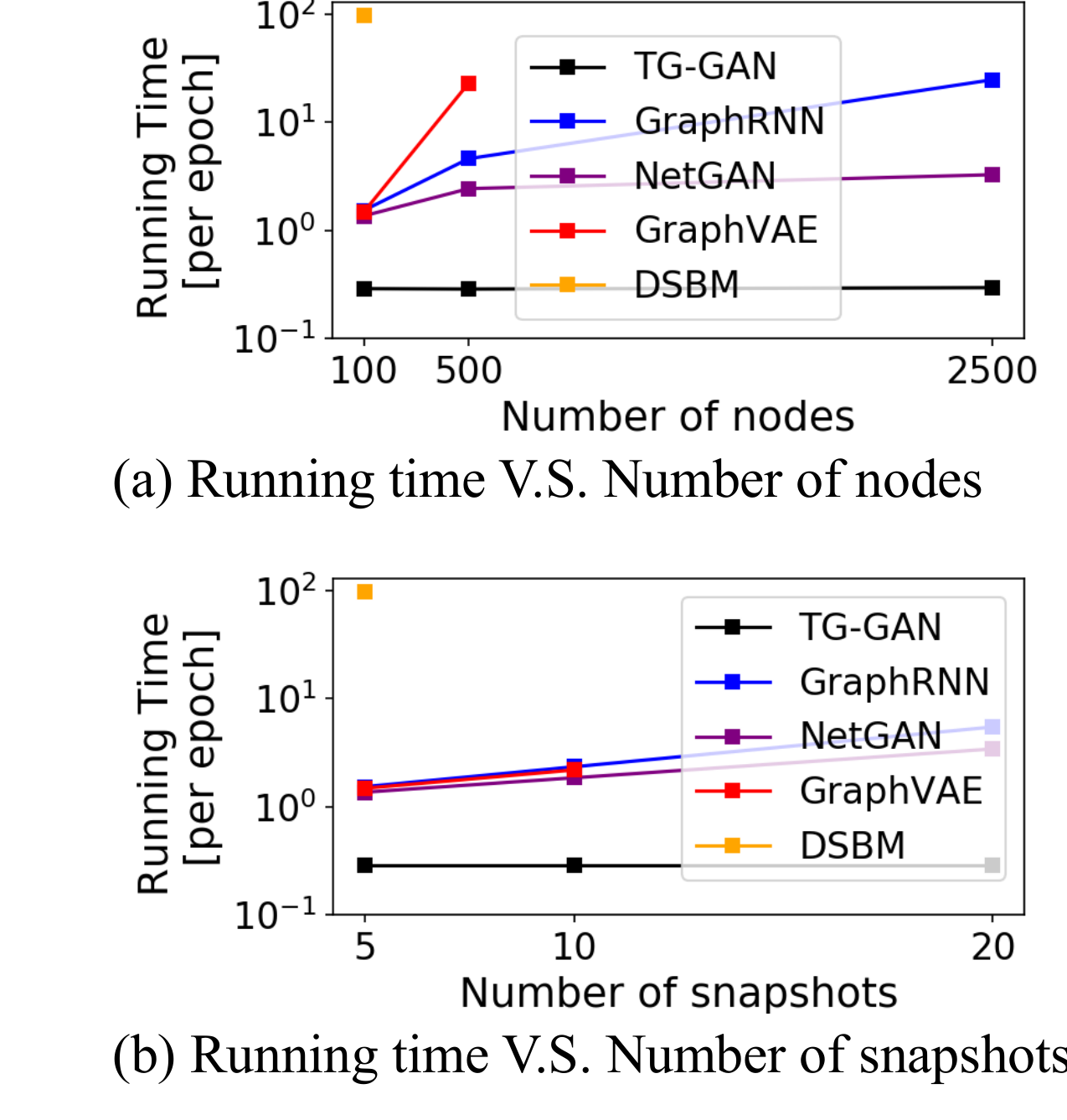}
      \captionof{figure}{Running time experiments}
      \label{fig:running_time}
  \end{minipage}%
  \hspace{-5mm}
  \begin{minipage}{0.7\linewidth}
  \centering
  \captionof{table}{MMD distances for discrete-time snapshot graph measures }
  \label{tab:mmd_discrete}
  \begin{tabular}{ m{.15in} | m{.4in} | m{.35in}  m{.25in}  m{.35in}  m{.35in}  m{.35in}  m{.25in}  m{.35in} }
    \toprule
    Nodes & \diagbox[width=.5in]{Method}{Metrics} & Betweenness Centrality & Broadcast Centrality & Burstiness Centrality & Closeness Centrality & Nodes' Temporal Correlation &  Receive Centrality & Temporal Correlation\\
    \midrule
    \multirow{4}{*}{Auth.} 
    & GraphRNN & 4.41e-06 & 0.3071 & 0.2090 & 0.0223 & 0.0057 & 0.3072 & 8.26e-06 \\
    & NetGAN & \textbf{4.40e-06} & 0.5996 & 0.0302 & 0.8119 & 0.0056 & 0.5971 & 8.27e-06 \\
    & GraphVAE & 4.41e-06 & 0.3999 & 0.1718 & 0.0390 & 0.0057 & 0.3973 & 8.26e-06 \\
    & DSBM & 0.9943 & 0.3598 & 0.0326 & 0.1016 & 0.0594 & 0.3628 & 0.0013 \\
    & TG-GAN & 5.05e-05 & \textbf{0.1468} & \textbf{0.0020} & 6.97e-04 & \textbf{0.0036} & \textbf{0.1365} & \textbf{5.23e-06} \\
    & Wenbin & 1.0 & 1.0568 & 0.0194 & \textbf{7.65e-07} & 1.9917 & 1.1609 & 0.3952 \\
    \hline
    \multirow{4}{*}{Metro} 
    & GraphRNN & 0.0815 & 0.7316 & 0.0031 & 2.54e-04 & NaN & 0.7351 & NaN \\
    & NetGAN & 0.0829 & 0.5946 & 0.0244 & 0.0166 & NaN & 0.5811 & NaN \\
    & GraphVAE & 0.0829 & 0.7509 & 0.0030 & 1.99e-04 & NaN & 0.7374 & NaN \\
    & DSBM & 0.7880 & 1.1403 & 0.0228 & 0.0164 & 0.0223 & 1.0444 & 2.01e-04 \\
    & TG-GAN & \textbf{0.0120} & \textbf{0.0257} & \textbf{0.0026} & \textbf{8.36e-06} & \textbf{2.86e-05} & \textbf{0.0266} & \textbf{4.95e-09} \\
    & Wenbin & 0.7880 & 1.5623 & 0.0241 & 1.79e-05 & 1.9943 & 1.5471 & 0.5078 \\
    \hline
    \multirow{4}{*}{100} 
    & GraphRNN & 0.9567 & \textbf{0.1658} & 0.3790 & 5.89e-04 & 0.0011 & 0.3023 & 1.81e-06 \\
    & NetGAN & 0.6497 & 0.7058 & 0.0092 & 0.2073 & 0.0014 & 0.2878 & \textbf{7.31e-07} \\
    & GraphVAE & 0.9567 & 0.2167 & 0.4138 & \textbf{5.37e-04} & 0.0011 & 0.3539 & 1.81e-06 \\
    & DSBM     & \textbf{0.0020} & 0.4016 & 0.0526 & 0.01666 & 0.0183 & \textbf{0.1317} & 1.33e-04 \\
    & TG-GAN    & 0.5606 & 0.2100 & \textbf{0.0026} & 0.0015 & \textbf{0.0010} & 0.2181 & 1.10e-06 \\
    \hline
    \multirow{3}{*}{500} 
    & GraphRNN & 0.7912 & 0.1556 & 0.1621 & - & 0.0049 & 0.4241 & 1.75e-07 \\
    & NetGAN & 0.7928 & 0.3253 & \textbf{0.0415} & - & 0.0049 & \textbf{0.0948} & 1.75e-07 \\
    & GraphVAE & 0.7928 & \textbf{0.0871} & 0.1921 & - & 0.0049 & 0.2858 & 1.75e-07 \\
    & TG-GAN    & \textbf{0.7231} & 0.2878 & 0.0842 & - &\textbf{ 0.0048} & 0.2087 & 1.75e-07 \\
    \hline
    \multirow{2}{*}{2500} 
    & GraphRNN & 0.8802 & 1.0239 & \textbf{9.65e-07} & - & 0.0044 & 1.2410 & 1.00e-08 \\
    & NetGAN & 0.8801 & 0.1200 & 0.0169 & - & 0.0044 & 0.0965 & 1.00e-08 \\
    & TG-GAN    & \textbf{0.8245} & \textbf{0.1549} & 0.4296 & - & \textbf{0.0043} & \textbf{0.1979} &  \textbf{9.76e-09} \\
  \bottomrule
  \end{tabular}
  \end{minipage}
\end{adjustwidth}
\end{figure*}

.
Table \ref{tab:real_continuous_measure} shows the mean values of distributions for some selected graph measures. The closer a value is to actual temporal graphs, the better the models are. We can see that TG-GAN is the closest to real datasets except Mean Degree. DSBM performs poorly in all cases.
Tables \ref{tab:mmd_continuous} and \ref{tab:mmd_discrete} demonstrate the effectiveness of the proposed TG-GAN framework for real-world datasets (identified as Auth. and Metro in the first column). The overall performance characteristics are comparable to the the synthetic datasets. Our analysis here focuses more on how effective TG-GAN is for different graph sparsities. The metro transport graph is an extremely sparse graph (typically 1-3 trips per day per traveler). We can see for the discrete-time measures in Table~\ref{tab:mmd_discrete} that GraphRNN and GraphVAE have NaN values for Node Temporal Correlation and Temporal Correlation, since they fail to model empty graphs for most temporal snapshots (no trips happened during that snapshot). This can be found in the qualitative visualization in Supplement \ref{supplement:qualitative}.


\begin{table*}
\scriptsize
\begin{adjustwidth}{0in}{-1.2in}
  \centering
  \begin{minipage}{0.3\linewidth}
    \centering
        \captionof{table}{Continuous-time measures}
        \label{tab:real_continuous_measure}
        \begin{tabular}{ m{.15in} | m{.4in} | m{.3in} m{.2in} m{.3in} m{.3in} }
        \toprule
        data & \diagbox[width=.5in]{Method}{Metrics} & Mean Degree & Average Group Size & Average Group Number & Mean Coordination Number\\
        \midrule
        \multirow{5}{*}{Auth.} 
        & GraphRNN     & 0.0205 & 1.1154 & 24.2373 & 0.5856  \\
        & GraphVAE     & \textbf{0.0204} & 1.1156 & 24.2360 & 0.5938  \\
        & DSBM         & 0.1918 & 0.9999 & 27.0000 & 2.2204  \\
        & TG-GAN        & 2.89e-05 & \textbf{1.0184} & \textbf{26.5156} & \textbf{0.0359} \\
        & Real & 0.0166 & 1.0276 & 26.2911 & 0.0959 \\
        \hline
        \multirow{5}{*}{Metro} 
        & GraphRNN     & 0.0026 & 1.0109 & 90.0176 & 0.0239 \\
        & GraphVAE     & 0.0026 & 1.0109 & 90.0160 & 0.0240 \\
        & DSBM         & 0.2544 & 0.9999 & 91.0000 & 2.22e-16 \\
        & TG-GAN        & \textbf{0.00077} & \textbf{1.0072} & \textbf{90.3523} & \textbf{0.01425}\\
        & Real & 0.00065 & 1.0077 & 90.3012 & 0.0154 \\
        \bottomrule
        \end{tabular}
  \end{minipage}%
  \hspace{-15mm}
  \begin{minipage}{0.7\linewidth}
    \centering
      \captionof{table}{MMD distances for continuous-time measures}
      \label{tab:mmd_continuous}
      \begin{tabular}{ m{.15in} | m{.4in} | m{.35in} m{.35in} m{.25in} m{.35in} m{.35in} m{.25in} m{.25in} }
        \toprule
        data & \diagbox[width=.5in]{Method}{Metrics} & Average Degree  & Mean Average Degree & Group Size & Average Group Size & Mean Coordination Number & Mean Group Number & Mean Group Duration\\
        \midrule
        \multirow{4}{*}{Auth.} 
        & GraphRNN     & 1.68e-05 & 1.42e-05 & 0.8053 & 0.0077 & 0.1829 & 0.9114 & \textbf{0.0654} \\
        & GraphVAE     & 1.70e-05 & 1.41e-05 & 0.8030 & 0.0077 & 0.1819 & 0.9083 & 0.0658 \\
        & DSBM         & 0.0002 & 0.0304 & 0.6344 & 0.0007 & 0.0087 & 0.2782 & 0.9315 \\
        & TG-GAN        & \textbf{3.38e-09} & \textbf{2.94e-09} & \textbf{0.1187} & \textbf{0.0004} & \textbf{0.0047} & \textbf{0.1035} & 0.1974 \\
        \hline
        \multirow{4}{*}{Metro} 
        & GraphRNN     & 5.92e-06 & 3.82e-06 & 0.1826 & 1.00e-05 & 7.31e-05 & 0.0745 & 1.0376 \\
        & GraphVAE     & 5.88e-06 & 3.84e-06 & 0.1831 & 1.02e-05 & 7.54e-05 & 0.0754 & 1.0320 \\
        & DSBM         & 0.0004 & 0.0634 & 1.2656 & 5.99e-05 & 0.0002 & 0.4198 & 0.8011 \\
        & TG-GAN        & \textbf{2.86e-08} & \textbf{1.45e-08} & \textbf{0.0065} & \textbf{2.92e-07} & \textbf{1.10e-06} & \textbf{0.0020} & \textbf{0.0910}\\
        \hline
        \multirow{4}{*}{100} 
        & GraphRNN     & 4.80e-05 & 8.16e-06 & 1.4152 & 0.0012 & 0.0019 & 1.4899 & 0.0342 \\
        & GraphVAE     & 7.72e-05 & 6.10e-06 & 1.4155 & 0.0012 & 0.0022 & 1.4933 & 0.0317 \\
        & DSBM         & 0.0083 & 0.0488 & \textbf{0.9932} & 1.8548 & 1.1320 & \textbf{1.0690} & \textbf{0.0020} \\
        & TG-GAN        & \textbf{6.46e-07} & \textbf{1.35e-06} & 1.24 & \textbf{0.0004} & \textbf{0.0005} & 1.3419 & 0.0041 \\
        \hline
        \multirow{3}{*}{500} 
        & GraphRNN     & 7.42e-06 & 3.03e-06 & 1.3557 & \textbf{0.0001} & \textbf{1.17e-05} & 1.4287 & 0.0615 \\
        & GraphVAE     & 6.96e-06 & 2.46e-06 & 1.2291 & 0.0002 & 0.0002 & 1.2982 & \textbf{0.0227} \\
        & TG-GAN        & \textbf{1.10e-06} & \textbf{2.10e-06 }& \textbf{0.8000} & 0.2312 & 1.8727 & \textbf{0.0800} & 0.2032 \\
        \hline
        \multirow{2}{*}{2500} 
        & GraphRNN     & 1.80e-06 & 9.33e-07 & \textbf{1.0961} & \textbf{0.0002} & \textbf{0.0002} & \textbf{1.1154} & 0.78380 \\
        & TG-GAN        & \textbf{4.78e-07} & \textbf{2.57e-07} & 1.1189 & \textbf{0.0002} & \textbf{0.0002} & 1.1292 & \textbf{0.0377} \\
      \bottomrule
      \end{tabular}
  \end{minipage}
\end{adjustwidth}
\end{table*}

\subsection{Qualitative analysis}
\label{sec:exp_qualitative}
We demonstrates a visualization (Figure \ref{fig:snapshot_viz}) for three datasets: user authentication graphs, metro transport graphs, and 100-node scale-free graphs. The y axis is the index of all nodes. x axis is index of time snapshots. The arc between two nodes is a temporal edge. The darker an edge is, the more graph samples it exists. Qualitative visualizations for all the other competing methods are given in Supplement\ref{supplement:qualitative}. It shows that TG-GAN indeedly capture the temporal patterns of real graphs.

\section{Conclusions}
\label{sec:conclusions}
To effectively model generative distributions of temporal graphs and retain continuous-time information, we propose the first-of-its-kind TG-GAN framework. It learns the representation of temporal graphs via temporal random walks, which includes a novel temporal generator to model truncated temporal random walks with profile information considering time dependency and  time constraint. A new temporal discriminator train the temporal generator with real training data from a novel walk sampler. Extensive experiments with synthetic and real-world datasets demonstrate advantages of TG-GAN model over existing deep and prescribed models.



\bibliographystyle{unsrt}
\bibliography{main}

\appendix

\section{Supplements}

\subsection{Temporal scale-free random graph simulation}
\label{supplement:simulation}

A directed scale-free graph \cite{bollobas2003directed} create a new edge from a new in-node, an existing node, or a new out-node by sampling a multinormial distribution of three probabilites $\langle \alpha, \beta, \gamma \rangle, \forall \alpha+\beta+\gamma =1$, which is adopted in Networkx \footnote{https://networkx.github.io/documentation/stable/index.html} library. We modify this edge generation procedure to a temporal dependent generation. The general idea is to append a continuous-time value to generated edge in each constructing step. First, a uniform distribution within $[0, 1]$ is used to sample a time value, $t \sim Uniform()$. And, we know that Unifrom distribution output value from $0$ to $1$. By comparing $t$ with cumulative probability $\langle \alpha, \alpha+\beta, 1 \rangle$, if $t \in [0, \alpha] $ a new in-node  indexed as $|V|+1$ is added and an exiting node is chosen from $V$ with probability $p(v = v_i)$ as an out-node, where $d_{in}$ is the function to get degree of $v_i$, $\delta_{in}$ is a hyper-parameter, $r$ is a uniformly random generated real number. A temporal edge is created for them with time-stamp $t$. If $t \in (\alpha, \alpha+\beta]$, two existing nodes $u_i$ and $v_i$ are chosen, and a temporal edge is created with time $t$. If $t \in (\alpha+\beta, 1]$, a new out-node is got, and a temporal edge to a chosen exiting node is created with time $t$. The choice of existing in-node $p(u = u_i) = \frac{d_{in}(u_i) + \delta_{in}}{|E| + \delta_{in} r}$, where $d_{in}$ is the function to get normalized in-degree of $u_i$, $\delta_{in}$ is a hyper-parameter, $r$ is a uniformly-random-generated real number. The choice of existing out-node $p(v = v_i) = \frac{d_{out}(v_i) + \delta_{out}}{|E| + \delta_{out} r}$, where $d_{out}$ is the function to get normalized out-degree of $v_i$, $\delta_{out}$ is a hyper-parameter, $r$ is another uniformly-random-generated real number. For more details, check \cite{bollobas2003directed}. After each constructing step, the ellapsed time is cumulated. This process is terminated untill either number of edges is equal to number of nodes, or a preseted maximum time range is reached. And, we borrow part of the codes from Networkx \footnote{https://networkx.github.io/documentation/stable/index.html} library`s original Scale-free graph codes.

\subsection{Competing methods details}
\label{supplement:revocer}

Modifications of adapting competing methods developed for static graphs to temporal graphs are described as follows:

\textbf{GraphRNN}. This is a recent state-of-the-art deep generative method. It is developed for a set of static graph samples. And, it is scalable to very large graphs. We use the default parameters provided in the GraphRNN code. 

\textbf{GraphVAE}. This is also a secent development deep generative method. It is also targeted for static graphs. Its complaxity analysis makes it only runnable for small graphs. The original paper do not have a published code, so the code in GraphRNN paper is used. Also, default parameters are used.

\textbf{DSBM}. It is most recent development of prescribed models for dynamic graphs based on Stochastic Blocks Models. It utilizes a Markovian transition to model the dynamic in temporal graphs.

Our models can adapt to start-time and end-time easily by adding additional LSTM cell for each temporal edge, if other applications need end-time evaluation. 
The performance of our new TG-GAN framework was compared with the above models. 
Another question in those three models is how to convert the snapshots back to continuous time. We simply choose the middle point of time in each snapshot as the real time of a generated temporal edge. Also, notice that temporal edges are modeled as a time point for all the datasets and generated data. For continuous-time measures, the existing of temporal edges have a start time and end time. For simplicity, we assume a constant time for all temporal edges to exist. This assumption is also true for real-world authentication graph which only has one timestamp for each edge. For transport graph, the existing time of a temporal edge is the travel time from one station to another station, which is almost fixed in metro schedules and relatively small compared to a whole 24 hours. It is also a reasonable assumption to use the same small edge existing time. 

\subsection{MMD}
In short, MMD measures how the distribution of one set of samples is similar to another set of samples. Given $\pmb{X} \in \mathbb{R}^{n \times k}$, each row $\pmb{X}_{i, }$ is a sampled vector from a unknown distribution. There is another $\pmb{X}' \in \mathbb{R}^{n' \times k}$, each row $\pmb{X}'_{i, }$ is a sampled vector from another unknown distribution. We have $MMD(\pmb{X}, \pmb{X}')$ as a distance measurement of these two sample sets. $MMD(\pmb{X}, \pmb{X}') = 0$ means two sets are exactly the same. In this experiment, different empirical graph measures are chosen for $\pmb{X}$. For example, the continuous-time average degree distribution of one graph sample is $\pmb{X}_{i,} \in \mathbb{R}^{1 \times |V|}$, where $|V|$ is the number of nodes. We can compute $MMD(\pmb{X}, \hat{\pmb{X}})$ to see if graphs samples generated from a trained model $\pmb{X}$ is close to real graph samples $\hat{\pmb{X}}$. 
\subsection{Parameter-tuning}
\label{supplement:tuning}
These includes a set of hyper-parameters that can be tuned for TG-GAN to achieve the best performance. Here are the list: learning rate [0.003, 0.003], generator node embedding size [node number / 2], discriminator node embedding size [node number / 2], $L2$ penalty of discriminator [$5e-5$], $L2$ penalty of generator [$1e-7$], up-project of $x, y$ [16-64], up-project of $t$ [32-128], up-project of node $v$ [32-128], generator LSTM cell state $c, h$ [$[100, 20], [50, 10], [100], [50]$], discriminator LSTM cell state $c, h$ [$[80, 20], [40, 10], [80], [40]$], start temperature of gumbel-max [5], wasserstein penalty [1, 10], time decoding methods [Gaussian, Gamma, Beta, Deep random time sampler], time constraint activation methods [Minmax, clipping, Relu], initial noise type [Uniform, Gaussian]. More potential hyper-parameter are released in github in the future.

\subsection{Additional experimental results analysis}
\label{supplement:qualitative}

\begin{figure*}
    \centering
        \includegraphics[width=\linewidth, trim={1cm 3cm 3.5cm 3cm},clip]{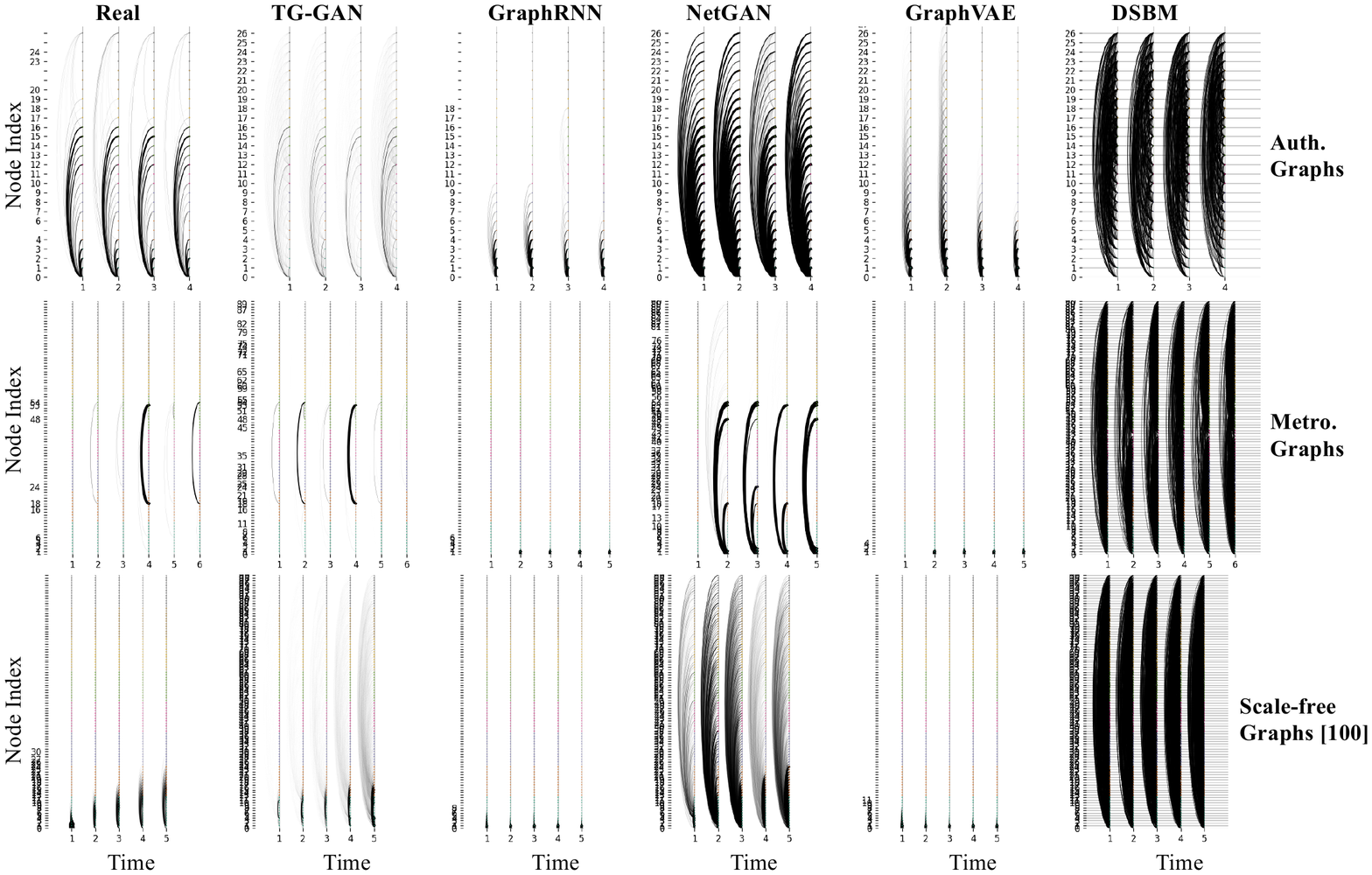}
        \captionof{figure}{Comparison of the real graphs (left), TG-GAN and four comparison methods for three different datasets}
        \label{fig:all_snapshot_viz}
\end{figure*}

Figure~\ref{fig:all_snapshot_viz} shows qualitative visualizations for the various methods and datasets. 
The small dots on each column represent nodes. The dots repeat across all temporal snapshots. The arc lines are edges. The figures are created by first, converting each graph sample to snapshots, and then, for each snapshot, summing up counts of an edge in all  graph samples. The more counts an edge receives, the darker the edge is shown. 

For user authentication graphs, we observe that this graph is densely connected in  sub-regions of the whole graph. TG-GAN mimics the real graph topology quite well. GraphRNN and GraphVAE show a trend towards similar topology. More training could potentiall result in better performance. DSBM does not create a graph close to  the actual topology.

For the metro transport graph,a good example of an extremely sparse graph, typically one edge has only one snapshot. Most of the temporal edges happen in one or two snapshots. We can see that TG-GAN is the only model to capture this challenging sparse graphs.

For 100-node scale-free synthetic graph, we can see that edges are connected increasingly as time grows. First, it did agree with typical scale-free graph pattern though it is a time-dependent growth. Then, TG-GAN performs the best to mimic this growing connection pattern. GraphRNN and GraphVAE also capture the overall trends. Probabily, more training would improve their performance. DSBM do not perform well too.

\end{document}